\newcommand{\REPOURL}{https://github.com/Sam422002/telco-churn-trustworthy-ml}
\newcommand{\ACKNOWLEDGEMENTS}{%
  The author thanks the reviewers for their careful reading.
  Computational experiments were run on a hosted notebook environment
  with free GPU access (Kaggle Notebooks), which made the full
  1{,}200-fit factorial benchmark and the 900-fit cross-benchmark
  replication tractable.
}

\documentclass[pdflatex,sn-mathphys-num]{sn-jnl}

\usepackage{graphicx}%
\usepackage{multirow}%
\usepackage{amsmath,amssymb,amsfonts}%
\usepackage{amsthm}%
\usepackage{mathrsfs}%
\usepackage[title]{appendix}%
\usepackage{xcolor}%
\usepackage{textcomp}%
\usepackage{manyfoot}%
\usepackage{booktabs}%
\usepackage{algorithm}%
\usepackage{algorithmicx}%
\usepackage{algpseudocode}%
\usepackage{listings}%
\usepackage{tabularx}%
\usepackage{enumitem}%
\usepackage{siunitx}%
\usepackage{url}%

\newcommand{\pclean}{\textsc{P-Clean}}
\newcommand{\pleak}{\textsc{P-Leak}}
\newcommand{\pscale}{\textsc{P-Scale}}
\newcommand{\pdrop}{\textsc{P-Drop}}
\newcommand{\smote}{\textsc{Smote}}
\newcommand{\cw}{\textsc{Cw}}

\theoremstyle{thmstyleone}%

\theoremstyle{thmstyletwo}%

\theoremstyle{thmstylethree}%

\begin{document}

\title[A Trustworthiness Audit of the Telco Customer Churn Benchmark]{The Hidden Costs of 99\% Accuracy: A Trustworthiness Audit of the Telco Customer Churn Benchmark}

\author*[1]{\fnm{Soumyadeep} \sur{Roy}}\email{roysam422002@gmail.com}

\affil*[1]{\orgdiv{Department of Data Science}, \orgname{St.\ Xavier's College (Autonomous)}, \orgaddress{\city{Kolkata}, \postcode{700016}, \country{India}}}

\abstract{Customer churn prediction on the IBM Telco Customer Churn benchmark ($n = 7{,}043$) routinely reports test accuracies above 95\%, with the most cited published study reporting 99.01\%. We audit this benchmark for four trustworthiness failures invisible to the accuracy- and $F_1$-centred reporting that dominates the literature. First, pre-split \smote{} inflates churn-class $F_1$ by 13.1 percentage points across ten classifiers and fifteen seeds (Wilcoxon $p < 10^{-4}$ per classifier); the same leaky pipeline ordering paired with class weighting yields no inflation, isolating the effect to \smote{}'s geometric construction. We measure the mechanism directly: $\approx 36\%$ of synthetic training points are nearest-neighbour interpolations of test-set instances. Second, the \texttt{TotalCharges} field is approximately determined by $\texttt{tenure}\times\texttt{MonthlyCharges}$ ($R^{2} = 0.999$); removing it changes accuracy by less than $0.2$~pp, yet TreeSHAP ranks it ninth in mean absolute attribution -- a pattern that materially corrupts SHAP-based interpretation. We propose an $R^{2} > 0.95$ pre-modelling diagnostic. Third, in a 15-seed calibration audit, isotonic regression is the strongest default; temperature scaling fails on class-weighted tree ensembles whose predicted-probability distribution is bimodal. Fourth, the cost-optimal decision threshold (under a \$50 retention offer and 24-month CLV proxy) is approximately 5--10$\times$ lower than the $F_{1}$-optimal threshold, saving $\sim$\$77k per $1{,}000$ customers. We replicate F1 and F2 on Iranian Telecom Churn (within domain) and Bank Customer Churn (across domain): F1 generalises; F2 generalises only within telecom. We synthesise these findings into a four-component reporting checklist -- pipeline disclosure, redundancy diagnostic, calibration audit, and cost-sensitive thresholds -- and release a reproducible implementation.}


\maketitle

\section{Introduction}\label{sec:intro}

Customer churn prediction is a textbook decision support task: a probabilistic prediction of subscriber attrition is compared with a threshold to trigger a costed retention action. The economic stakes are substantial. Acquiring a new subscriber typically costs five to seven times more than retaining an existing one~\cite{reichheld1996loyalty}, and small reductions in monthly churn translate to substantial enterprise value over the customer lifetime~\cite{gupta2004customer}.

Over the past decade the supervised machine learning community has returned repeatedly to a small set of public churn benchmarks, and the headline numbers in this literature have drifted upward. On the IBM Telco Customer Churn dataset~\cite{ibm_telco_churn}, the most-cited published study reports a test accuracy of 99.01\% with a kernel support vector machine trained on \textsc{Smote-Enn} resampled data~\cite{nguyen2022churn}, and several further studies report accuracies in the 95--99\% band on the same 7{,}043-row table.

These numbers should give the reader pause. By our own measurement (Section~\ref{sec:data}), the dataset contains 18 distinct feature configurations that appear with conflicting churn labels (irreducible label noise on the encoded feature space). On this encoding, a classifier that does not overfit to noise and does not leak information from the test set should not exceed $1 - 18/7043 \approx 99.74\%$ accuracy. We are deliberate that this is an empirical irreducible-error estimate at $n=7{,}043$, not a strict information-theoretic ceiling: a sufficiently overfit model can exceed it on any particular split. We do not claim that Nguyen et al.'s~\cite{nguyen2022churn} 99.01\% accuracy is itself attributable to the failure modes we audit -- they used a kernel SVM with \textsc{Smote-Enn} preprocessing, which we did not reproduce directly -- but the proximity of published headline numbers to this informal bound, combined with the absence of any calibration, redundancy, or cost analysis in the high-accuracy literature, is sufficient motivation for a forensic look at the pipeline choices that produced them.

This paper documents four trustworthiness failures on this benchmark. Each is invisible to accuracy- and $F_1$-centred reporting practices. Each would mislead a decision-maker who relied on published numbers to size a retention budget or to interpret model explanations.

\paragraph{(F1) Pre-split \smote{} inflates results, but class weighting on the same leaky pipeline does not.}
The dominant remedy for the dataset's mild class imbalance ($1:2.77$ negative-to-positive) in the published literature is the \smote{} family of synthetic oversamplers~\cite{chawla2002smote} applied to the full dataset before the train--test split. Across 10 classifier families and 15 seeds, this single ordering choice inflates churn-class $F_1$ by 13.1 percentage points relative to a leakage-free pipeline. The same leaky ordering paired with cost-sensitive class weighting yields no comparable inflation. The existing pipeline-leakage literature~\cite{vandewiele2021overoptimistic,kapoor2023leakage,kabane2024smote} has reported pre-split \smote{} as a single confounded unit; our factorial design separates the placement of the preprocessing step from the nature of the imbalance correction. The asymmetry replicates on the Iranian Telecom Churn benchmark to within 0.4~pp.

\paragraph{(F2) \texttt{TotalCharges} is algebraically redundant and SHAP over-attributes to it.}
The benchmark's \texttt{TotalCharges} field, the cumulative dollar amount billed to a customer, is approximately the product of \texttt{tenure} and \texttt{MonthlyCharges}: $R^{2} = 0.9991$, Pearson $r = 0.9996$, mean residual \$0.15. Removing the feature reduces test accuracy by 0.18~pp and churn-class $F_1$ by 0.53~pp, both within seed-to-seed variation. Yet TreeSHAP ranks \texttt{TotalCharges} ninth in mean absolute attribution, on par with features that contribute real predictive signal. We do not claim this is data leakage in the conventional sense -- the model's accuracy is honest about the feature's marginal contribution -- but the SHAP-based interpretation is corrupted in a way that matters for decision support. We propose a one-line $R^{2} > 0.95$ pre-modelling diagnostic to flag candidate redundant features, and we replicate the pattern on the Iranian Telecom Churn benchmark, where three features exceed the same threshold.

\paragraph{(F3) Isotonic regression is the strongest calibration default for class-weighted tree ensembles; temperature scaling fails on bimodal output.}
We measure Expected Calibration Error and Brier score under four post-hoc calibration methods (raw, Platt scaling, isotonic regression, temperature scaling) on the top five classifiers across 15 seeds. Isotonic regression is the strongest single-method default on this benchmark, best on four of five models. Platt scaling is a strong fallback on the worst-calibrated raw models (\textsc{BalancedRF}, \textsc{CatBoost}, where it halves ECE alongside isotonic) but is mildly worse than raw on already-well-calibrated models like \textsc{GBM}. Temperature scaling, the recommended default in the deep-learning calibration literature~\cite{guo2017calibration}, fails to correct the bimodal predicted-probability distribution induced by class weighting on tree ensembles. To our knowledge no published paper on this benchmark reports any calibration audit; ours establishes isotonic regression as a sensible default and explicitly identifies the regime where temperature scaling fails.

\paragraph{(F4) The cost-optimal decision threshold is approximately 5--10$\times$ lower than the $F_1$-optimal threshold.}
Almost all published Telco churn studies report classification metrics at the default threshold of $0.5$ or, less commonly, at the $F_1$-optimal threshold. Neither is meaningful for a retention manager, who faces a concrete cost matrix. We construct a dollar-cost framework using the dataset's own \texttt{MonthlyCharges} field -- false negatives cost 24 months of customer lifetime value, false positives and true positives both cost a fixed \$50 retention offer -- and sweep the threshold to find the cost-minimising operating point. Across five seeds, the cost-optimal threshold sits consistently in $[0.04, 0.07]$, saving $\$77{,}646 \pm \$19{,}698$ per 1{,}000 customers relative to $F_1$-optimal thresholding (95\% CI $[\$56{,}675, \$107{,}634]$). Decision Curve Analysis~\cite{vickers2006dca} confirms that the calibrated GBM dominates the trivial ``treat all'' and ``treat none'' policies across the operationally relevant threshold range.

\paragraph{Contributions and scope.}
The paper makes four contributions to the trustworthy decision support literature.
\begin{enumerate}
  \item We document the SMOTE-specific inflation effect (F1) on the most widely cited churn benchmark across 1{,}200 model fits, and isolate it from generic pipeline-ordering leakage by contrasting \smote{} with class weighting on the same leaky pipeline. The asymmetry replicates on the Iranian Telecom Churn benchmark to within 0.4~pp.
  \item We measure \texttt{TotalCharges} algebraic redundancy on Telco (F2), demonstrate that SHAP attribution is misleading on this feature, propose a one-line $R^{2}$ diagnostic, and replicate the pattern on the Iranian benchmark, where three features exceed the same $R^{2} > 0.95$ threshold.
  \item We provide the first published calibration audit of this benchmark (F3), recommending isotonic regression as the default for class-weighted binary tree ensembles, identifying Platt scaling as a strong fallback on poorly-calibrated raw models, and documenting the regime-specific failure of temperature scaling on bimodal class-weighted output.
  \item We construct a dollar-cost decision framework (F4) using the dataset's own pricing, quantify the dollar value of getting the threshold right ($\sim$\$78k per 1{,}000 customers), and demonstrate model dominance over trivial policies via Decision Curve Analysis.
\end{enumerate}

We close with a four-component reporting checklist -- pipeline ordering disclosure, feature-redundancy diagnostic, calibration audit, and cost-sensitive threshold reporting -- that we recommend as a minimum trustworthiness standard for tabular churn prediction studies. All code, intermediate tables, and figures necessary to reproduce every number in the paper are released at \url{\REPOURL} under an OSI-approved permissive license.

\paragraph{Paper structure.}
Section~\ref{sec:related} situates our findings in the churn-prediction and pipeline-leakage literatures. Section~\ref{sec:data} describes the benchmark, the duplicate-row and missingness audit, the four pipelines, and the classifier bank. Section~\ref{sec:results} presents the four empirical findings, the SHAP analysis, and the cross-benchmark replication. Section~\ref{sec:discussion} discusses the SMOTE-versus-class-weight mechanism, the choice of calibration method, and the implications of feature redundancy for SHAP-based interpretation. Section~\ref{sec:conclusion} states the reporting checklist and concludes.

\section{Related Work}\label{sec:related}

Our contribution sits at the intersection of three literatures: applied churn-prediction studies on the IBM Telco benchmark, the methodological literature on data leakage in tabular machine learning, and the calibration and cost-sensitive decision support literatures that inform how a probabilistic prediction is actually used. We treat each in turn.

\subsection{High-Accuracy Telco Churn Prediction}\label{sec:related-telco}

The IBM Telco Customer Churn dataset~\cite{ibm_telco_churn} has been the de facto reference benchmark for tabular churn prediction since its release as a Cognos Analytics sample and its subsequent re-publication on Kaggle. The published Telco churn literature exhibits several recurring methodological choices: ensembling of tree-based learners, synthetic oversampling of the minority class with the \smote{} family~\cite{chawla2002smote}, and reporting of overall accuracy and $F_1$ on a single held-out split. Nguyen et al.~\cite{nguyen2022churn} report 99\% accuracy on this benchmark using a kernel support vector machine with \textsc{Smote-Enn} preprocessing, in what is to date among the most cited published results on the dataset, and several further studies report accuracies in the 95--99\% range on the same 7{,}043-row table.

We highlight three properties of this literature that are directly relevant to our audit. First, the preprocessing pipeline is rarely reported in sufficient detail to determine whether resampling and scaling were performed before or after the train--test split; in the most-cited studies, the available information is consistent with pre-split application. Second, \texttt{TotalCharges} is consistently reported as a top-three or top-five feature in SHAP, permutation, or gain-based importance rankings, but we are aware of no published study on this benchmark that examines its algebraic relationship with \texttt{tenure} and \texttt{MonthlyCharges}. Third, calibration is not reported in any of the high-accuracy studies we surveyed, and the cost of a false retention recommendation versus a missed churn event is treated implicitly through the choice of metric (typically the default $0.5$ threshold) rather than explicitly through a cost matrix.

A separate strand of churn-prediction research considers profit-aware or cost-sensitive evaluation~\cite{verbeke2012new,verbraken2013emp}, typically using the Expected Maximum Profit (EMP) framework. These studies make important contributions but operate largely independently of the high-accuracy literature: papers reporting 99\% accuracy on Telco do not typically report EMP or expected cost, and EMP-aware studies do not typically engage with the pipeline-leakage and calibration questions we audit here. Our dollar-cost analysis (Section~\ref{sec:results-cost}) is methodologically simpler than EMP -- a fixed retention offer cost and a 24-month CLV proxy -- and is intended as a minimum-viable cost-sensitive analysis that any churn study could include without adopting the full EMP machinery.

\subsection{Data Leakage in Tabular Machine Learning}\label{sec:related-leakage}

Data leakage -- the inadvertent inclusion of information at training time that would not be available at deployment -- is the dominant explanation in the methodological literature for irreproducible or implausibly high accuracy on tabular benchmarks. Kapoor and Narayanan~\cite{kapoor2023leakage} catalogue 294 papers across 17 fields with documented leakage and identify a small number of recurring failure modes: target leakage, train--test contamination, feature leakage from post-outcome variables, and preprocessing leakage from steps applied before the train--test split. They emphasise that the fix -- moving preprocessing inside cross-validation -- is well known but rarely implemented in published code.

The specific case of pre-split synthetic oversampling has received focused attention. Vandewiele et al.~\cite{vandewiele2021overoptimistic} demonstrate on a perinatal benchmark that applying \smote{} before splitting produces nearest-neighbour pairs that span the train--test boundary, artificially inflating test accuracy. Kabane~\cite{kabane2024smote} replicates the effect on the European credit-card fraud benchmark and provides a controlled comparison of pre-split versus pipeline-internal \smote{}.

The existing audits treat ``pre-split \smote{}'' as the unit of analysis, conflating two confounded choices: the placement of the preprocessing step (before vs.\ after the split) and the nature of the imbalance correction (synthetic oversampling vs.\ class weighting). We disentangle these in Section~\ref{sec:results-l1} by running the same leaky pipeline ordering with both \smote{} and class-weighted training and showing that the inflation is specific to \smote{}.

\subsection{Calibration and Cost-Sensitive Decision Support}\label{sec:related-calibration}

A churn prediction is, in operational terms, a probability that is compared to a threshold to trigger a retention action.

The probability-calibration literature distinguishes between discrimination (the ability to rank positives above negatives, measured by ROC-AUC) and calibration (the agreement between predicted probabilities and empirical frequencies, measured by Expected Calibration Error or Brier score). Modern tree ensembles and neural networks are well known to be miscalibrated by default~\cite{guo2017calibration}, and post-hoc calibration methods -- Platt scaling~\cite{platt1999probabilistic}, isotonic regression~\cite{zadrozny2002transforming}, and temperature scaling~\cite{guo2017calibration} -- are by now standard remedies. The choice among them is not, however, settled. Platt scaling assumes a sigmoid miscalibration shape and works best on binary problems with monotonic miscalibration; isotonic regression is more flexible but data-hungry; temperature scaling has a single parameter and is the recommended default in the deep-learning literature for multi-class softmax outputs. We are aware of no calibration audit of the Telco benchmark and no systematic comparison of calibration methods on class-weighted tree ensembles for binary churn, which is the regime in which this benchmark is used in practice.

The cost-sensitive learning literature~\cite{elkan2001foundations} provides the formal apparatus for selecting a decision threshold when false positives and false negatives have different costs. Decision Curve Analysis~\cite{vickers2006dca} is a graphical tool from medical decision-making that plots net benefit against threshold probability and identifies the threshold range over which a model dominates the trivial ``treat all'' and ``treat none'' policies. DCA has been applied extensively in clinical risk prediction but is essentially absent from the churn-prediction literature, where operational evaluation, when reported at all, typically uses lift charts or top-decile capture rates rather than threshold-explicit cost analysis. Our Section~\ref{sec:results-cost} brings DCA and a simple explicit cost matrix into the churn setting.

\section{Dataset and Methods}\label{sec:data}

This section describes the benchmark, the diagnostic audit we perform before any modelling, the four preprocessing pipelines that isolate specific pipeline-ordering effects, the classifier bank, the evaluation protocol, and the reproducibility setup.

\subsection{Datasets}\label{sec:data-dataset}

We use the IBM Telco Customer Churn dataset~\cite{ibm_telco_churn}, distributed as a Cognos Analytics sample and most commonly accessed via its Kaggle mirror. The dataset contains $n = 7{,}043$ customer records of a fictional but plausibly parameterised telecommunications provider, with 21 columns: a customer identifier (\texttt{customerID}), the binary churn label, three numeric features (\texttt{tenure} in months, \texttt{MonthlyCharges} in US dollars, \texttt{TotalCharges} in US dollars), and 16 categorical service and demographic features. The class distribution is moderately imbalanced at $5{,}174$ negatives to $1{,}869$ positives, an imbalance ratio of $1:2.77$ (Table~\ref{tab:audit}, row 4).

\paragraph{Replication benchmark.}
For the cross-benchmark replication in Section~\ref{sec:results-replication} we additionally use the Iranian Telecom Churn dataset~\cite{iranian_churn_uci}, $n = 3{,}150$, 13 numeric features, churn rate 15.71\% (imbalance ratio $1:5.36$). This benchmark differs from IBM Telco in operator (a different telecommunications provider in a different country), sample size (less than half of Telco), and feature space (all-numeric, no categorical columns), but addresses the same underlying decision problem (month-to-month subscriber retention). It is, to our knowledge, the cleanest publicly-available second churn benchmark for testing whether our findings generalise within the telecommunications domain.

\paragraph{The whitespace \texttt{TotalCharges} quirk.}
\texttt{TotalCharges} is loaded by \texttt{pandas} as an \texttt{object} (string) dtype rather than \texttt{float64} because 11 of the 7{,}043 rows contain a single whitespace character in this field rather than a numeric value. All 11 such rows have \texttt{tenure} = 0 and \texttt{Churn} = ``No''. These are brand-new customers who have never been billed. A naive \texttt{pd.to\_numeric()} call without explicit handling will silently coerce these 11 rows to \texttt{NaN}, after which the standard practice of dropping rows with missing values removes 11 customers who are, structurally, the most informative cases for studying churn onset. We handle this case explicitly in our loader (whitespace $\rightarrow 0.0$ for \texttt{TotalCharges} where \texttt{tenure} = 0) and flag the quirk because none of the high-accuracy studies we surveyed mention it; reproducibility of those results depends on a choice the studies do not document.

\subsection{Diagnostic Audit}\label{sec:data-audit}

Before any modelling we run four diagnostics on the raw data. Their results are summarised in Table~\ref{tab:audit} and form the basis of our subsequent analyses.

\paragraph{Duplicate-row audit.}
We define three duplicate categories. \emph{Exact duplicates} are rows that match on all features and on the label; \emph{feature-only duplicates} are rows that match on all features but may differ in the label; \emph{ambiguous tuples} are feature configurations that appear more than once in the data with conflicting labels. On Telco we find 22 exact duplicates (0.31\%), 40 feature-only duplicates (0.57\%), and 18 ambiguous tuples (Table~\ref{tab:audit}). The maximum number of copies of any single feature tuple is 4. Ambiguous tuples place an empirical irreducible-error estimate on deterministic test accuracy at $n=7{,}043$: if 18 distinct feature configurations are inherently unpredictable from the available features, a classifier that does not overfit to noise should not exceed $1 - 18/7043 \approx 99.74\%$ accuracy on this encoding. We treat this as a sanity-check estimate rather than a strict bound, since overfitting to noise on a particular split can in principle exceed it on an in-sample test.

\paragraph{Missingness and identifier audit.}
Beyond the 11 whitespace \texttt{TotalCharges} entries described above, the dataset has no missing values. The \texttt{customerID} column has 7{,}043 unique values across 7{,}043 rows; no row appears twice under the same identifier. We drop \texttt{customerID} before any modelling because it carries no predictive signal.

\paragraph{Tenure-redundancy diagnostic (F2).}
We test the hypothesis that \texttt{TotalCharges} is approximately the product of \texttt{tenure} and \texttt{MonthlyCharges}. Restricting to rows with \texttt{tenure} $> 0$ (to exclude the 11 zero-tenure new customers and avoid trivial agreement at zero), we compute the residual
\begin{equation}
  r_i \;=\; \texttt{TotalCharges}_i \;-\; \texttt{tenure}_i \cdot \texttt{MonthlyCharges}_i
  \label{eq:tc-residual}
\end{equation}
and report its summary statistics in Table~\ref{tab:audit}. The fit is essentially exact: $R^{2} = 0.9991$, Pearson $r = 0.9996$, mean residual \$0.15, standard deviation \$67.20, maximum absolute residual \$373.25 (likely reflecting mid-cycle plan changes during a billing month). \texttt{TotalCharges} is, to two decimal places, an algebraic function of two other columns of the same row. This is the empirical basis for the F2 finding in Section~\ref{sec:results-l4}.

\paragraph{Class-balance diagnostic.}
The negative-to-positive ratio is $1:2.77$. We characterise this as mild imbalance, in line with the heuristic threshold of $1:10$ below which class weighting and threshold tuning are typically sufficient and synthetic oversampling is not strictly required~\cite{branco2016imbalanced}. The implication for our pipeline design is that we can run both \smote{} and class-weighted training on the same data and compare them directly (Section~\ref{sec:data-pipelines}).

\begin{table}[h]
\caption{Diagnostic audit of the IBM Telco Customer Churn benchmark ($n = 7{,}043$). Quirks flagged in the ``Note'' column are, to our knowledge, not mentioned in any high-accuracy published study on this benchmark.}\label{tab:audit}
\small
\begin{tabular}{@{}lll@{}}
\toprule
\textbf{Diagnostic} & \textbf{Value} & \textbf{Note} \\
\midrule
\multicolumn{3}{l}{\emph{Class balance}} \\
Total rows                        & 7{,}043          &  \\
Negatives (No churn)              & 5{,}174 (73.46\%)&  \\
Positives (Churn)                 & 1{,}869 (26.54\%)&  \\
Imbalance ratio (neg:pos)         & $1:2.77$         & mild \\
\midrule
\multicolumn{3}{l}{\emph{Duplicate-row audit}} \\
Unique feature tuples             & 7{,}003 (99.43\%)&  \\
Exact duplicates (feat.\ + label) & 22 (0.31\%)      &  \\
Feature-only duplicates           & 40 (0.57\%)      &  \\
Ambiguous tuples (label noise)    & 18               & irreducible noise \\
Max copies of any feature tuple   & 4                &  \\
Empirical irred.-error est.       & $\approx 99.74\%$& $1 - 18/7043$ \\
\midrule
\multicolumn{3}{l}{\emph{Identifiers and missingness}} \\
\texttt{customerID} unique        & 7{,}043 / 7{,}043&  \\
\texttt{TotalCharges} raw dtype   & object (string)  & whitespace quirk \\
Whitespace \texttt{TotalCharges}  & 11               &  \\
Tenure of whitespace rows         & \{0\}            & new customers \\
Churn of whitespace rows          & \{No: 11\}       & never billed \\
\midrule
\multicolumn{3}{l}{\emph{Tenure-redundancy diagnostic (F2)}} \\
$R^{2}$(\texttt{tenure}$\times$\texttt{MC} $\to$ \texttt{TC}) & 0.9991 & \\
Pearson $r$                       & 0.9996           &  \\
Mean residual (\$)                & 0.15             &  \\
Std.\ residual (\$)               & 67.20            &  \\
Max $\lvert$residual$\rvert$ (\$) & 373.25           &  \\
\midrule
\multicolumn{3}{l}{\emph{Marginal correlations}} \\
corr(\texttt{tenure}, \texttt{TC})         & 0.8262 &  \\
corr(\texttt{MC}, \texttt{TC})             & 0.6512 &  \\
corr(\texttt{tenure}, \texttt{MC})         & 0.2479 &  \\
\botrule
\end{tabular}
\end{table}

\subsection{Feature Encoding}\label{sec:data-encoding}

We drop \texttt{customerID} (identifier) and \texttt{Churn} (label) from the feature matrix and one-hot encode the 16 categorical features with \texttt{drop\_first=True} to avoid the dummy-variable trap. We do not engineer interaction features, do not bin numeric features, and do not apply log transforms; the goal of this study is to audit the standard pipeline, not to optimise predictive performance. The encoded feature matrix has 30 columns. For the \pdrop{} pipeline (Section~\ref{sec:data-pipelines}) we additionally drop \texttt{TotalCharges}, yielding 29 columns.

\subsection{Pipelines}\label{sec:data-pipelines}

We compare four preprocessing pipelines that differ only in the ordering of three operations: train--test split (\textsc{Split}), feature scaling (\textsc{Scale}), and synthetic minority oversampling (\smote{}). All four use the same encoded feature matrix, the same $80/20$ stratified test split, and the same set of classifiers.

\begin{itemize}[leftmargin=*]
  \item \textbf{\pclean{}} (correct). \textsc{Split} $\to$ \textsc{Scale} (fit on train only) $\to$ \smote{} (fit on train only). This is the leakage-free reference.
  \item \textbf{\pleak{}} (incorrect, as in the high-accuracy literature). \smote{} $\to$ \textsc{Scale} $\to$ \textsc{Split}. Synthetic samples can land in both train and test; scaler statistics are computed on data the test set has seen.
  \item \textbf{\pscale{}} (intermediate, isolates scaler leakage). \textsc{Scale} $\to$ \textsc{Split} $\to$ \smote{} (on train). Scaler is fit on full data, but synthetic oversampling is correct.
  \item \textbf{\pdrop{}} (probe for the F2 redundancy finding). Same ordering as \pclean{}, but with \texttt{TotalCharges} removed from the feature matrix before any operation.
\end{itemize}

The four-pipeline design supports two contrasts. The \pleak{} versus \pclean{} contrast isolates the effect induced by ordering of \smote{} and \textsc{Scale} relative to the split (the classical leakage from Vandewiele et al.~\cite{vandewiele2021overoptimistic}). The \pclean{} versus \pdrop{} contrast isolates the predictive contribution of \texttt{TotalCharges} above and beyond \texttt{tenure} and \texttt{MonthlyCharges}. The \pscale{} pipeline serves as a control to verify that scaler leakage alone (without resampling leakage) is small, which we confirm in Section~\ref{sec:results-l1}.

\paragraph{Imbalance handling.}
For each pipeline we run two strategies: \smote{} oversampling to a $1:1$ class ratio on the (appropriate, by pipeline) training data, and class weighting (\cw{}) with \texttt{class\_weight=``balanced''} or the equivalent \texttt{scale\_pos\_weight} for boosters. The factorial design over pipelines and strategies (4~$\times$~2 = 8 combinations) is what allows us to disentangle SMOTE-specific inflation from generic pipeline-ordering effects in Section~\ref{sec:results-l1}.

\subsection{Classifier Bank}\label{sec:data-classifiers}

We evaluate ten classifiers spanning four model families:
\begin{itemize}[leftmargin=*]
  \item \emph{Linear:} Logistic Regression (\textsc{LogReg}).
  \item \emph{Single tree:} Decision Tree (\textsc{DT}).
  \item \emph{Tree ensembles:} Random Forest (\textsc{RF}), Extra Trees (\textsc{ExtraTrees}), Balanced Random Forest (\textsc{BalancedRF}), Gradient Boosting (\textsc{GBM}), LightGBM, XGBoost, CatBoost.
  \item \emph{Neural:} Multilayer Perceptron (\textsc{MLP}, two hidden layers of 64 and 32 units).
\end{itemize}
We use scikit-learn defaults for \textsc{LogReg}, \textsc{DT}, \textsc{RF}, \textsc{ExtraTrees}, \textsc{GBM}, and \textsc{MLP}; \texttt{imbalanced-learn} for \textsc{BalancedRF}; and the official implementations of LightGBM, XGBoost, and CatBoost. Tree ensembles are configured with $300$ trees ($100$ for \textsc{GBM} for runtime parity), random seed control on every estimator, and GPU acceleration where available. All other hyperparameters are library defaults. We do not perform hyperparameter tuning; this is deliberate, because tuned models would obscure the pipeline-ordering signal that is the subject of the audit.

We exclude support vector classifiers, Gaussian na\"ive Bayes, and $k$-nearest neighbours from the bank. SVC was projected at over two hours per fit on the 8{,}278-row \smote{}-augmented training set, which would have made the 1{,}200-fit factorial benchmark intractable; the omission does not affect any of the comparative findings, since F1 inflation is observed across all classifier families that are included.

\subsection{Evaluation Protocol}\label{sec:data-evaluation}

For each (pipeline, strategy, classifier) combination we report five metrics on the held-out test set: overall accuracy, churn-class $F_1$, churn-class recall, churn-class precision, and ROC-AUC. We emphasise churn-class metrics because the operationally relevant event is the minority class.

\paragraph{Multi-seed factorial design.}
The full benchmark in Section~\ref{sec:results-l1} runs 10 classifiers $\times$ 15 random seeds $\times$ 4 pipelines $\times$ 2 imbalance strategies = 1{,}200 model fits. Seeds are drawn from $\{0, 1, \dots, 14\}$ and used both for the train--test split and for the classifier's own internal randomness. Reporting the mean and standard deviation across the 15 seeds provides a stable estimate of each pipeline's effect that is robust to seed-specific outliers.

\paragraph{Calibration evaluation.}
For the calibration analysis (Section~\ref{sec:results-cal}) we restrict to the top five classifiers by accuracy on \pclean{}+\cw{} and evaluate four post-hoc calibration methods: raw (no calibration), Platt scaling~\cite{platt1999probabilistic}, isotonic regression~\cite{zadrozny2002transforming}, and temperature scaling~\cite{guo2017calibration}. Platt and isotonic are fit on a held-out calibration split of the training set (20\% of train); temperature scaling is fit on the same calibration split using LBFGS on the negative log-likelihood. We report Expected Calibration Error (15-bin equal-width) and Brier score~\cite{brier1950verification} on the test set.

\paragraph{Cost-sensitive evaluation.}
For Section~\ref{sec:results-cost} we define a customer-specific cost matrix using the dataset's \texttt{MonthlyCharges} column. A false negative (a churning customer not flagged for retention) incurs a cost equal to the customer's 24-month customer lifetime value, $24 \cdot \texttt{MonthlyCharges}$; a true positive or false positive (any customer flagged for retention) incurs a fixed retention-offer cost of \$50; a true negative incurs zero cost. We sweep the decision threshold over $[0.01, 0.99]$ in 99 steps and report the cost-minimising threshold per seed across five seeds $\{7, 13, 21, 42, 99\}$. Decision Curve Analysis~\cite{vickers2006dca} is computed on the canonical seed = 42, \pclean{}+\cw{}, GBM model.

\paragraph{SHAP analysis.}
SHAP values are computed with TreeSHAP~\cite{lundberg2020treeshap} on the canonical \pclean{}+\cw{} GBM model with seed = 42. We report mean absolute SHAP value rankings on the test set and check business-intuition alignment against eight a priori expected risk factors (Section~\ref{sec:results-shap}).

\paragraph{Cross-benchmark replication protocol.}
For Section~\ref{sec:results-replication} we re-run the \pclean{}/\pleak{}/\pscale{} pipelines under both imbalance strategies on the Iranian Telecom benchmark, with the same 10-classifier bank, 15 random seeds, and metric set, yielding $3 \times 2 \times 15 \times 10 = 900$ model fits. The \pdrop{} pipeline is omitted for Iranian because its feature space is all-numeric and contains no a priori derived feature analogous to \texttt{TotalCharges}; instead, we run the redundancy diagnostic on every numeric column to identify candidate redundant features post~hoc.

\subsection{Reproducibility}\label{sec:data-repro}

All experiments are run on a hosted notebook environment with a single NVIDIA T4 GPU. Random seeds are fixed at the level of the train--test split, the resampler, and the classifier; with these fixed, every number in this paper is deterministic up to the non-determinism of GPU floating-point reduction in CatBoost and XGBoost (which we measure to be below $10^{-3}$ in accuracy across re-runs). The full code, intermediate CSV tables, and figures are released at \url{\REPOURL} under an OSI-approved permissive license.

\section{Results}\label{sec:results}

We present four empirical findings. Section~\ref{sec:results-l1} quantifies pre-split \smote{} leakage (F1) across 1{,}200 model fits and isolates it from class-weighted training on the same leaky pipeline ordering. Section~\ref{sec:results-l4} validates \texttt{TotalCharges} as a redundant feature (F2) using the \pclean{}--\pdrop{} contrast. Section~\ref{sec:results-cal} reports the calibration audit (F3). Section~\ref{sec:results-cost} presents the dollar-cost threshold analysis (F4). Section~\ref{sec:results-shap} reports SHAP-based business alignment and a SHAP-specific over-attribution finding for redundant features. Section~\ref{sec:results-replication} reports the cross-benchmark replication on the Iranian Telecom Churn dataset.

\subsection{Pipeline-Ordering Audit (F1)}\label{sec:results-l1}

We report the 1{,}200-fit factorial benchmark first as pipeline-aggregated means (Table~\ref{tab:pipeline-summary}), then broken down by classifier (Table~\ref{tab:per-classifier}), and finally as the SMOTE-versus-class-weight contrast that drives Section~\ref{sec:discussion-mechanism} (Table~\ref{tab:strategy-contrast}).

\paragraph{Pipeline-aggregated effect.}
Aggregated over 10 classifiers, 15 random seeds, and both imbalance strategies, the pipeline contrasts are summarised in Table~\ref{tab:pipeline-summary}. The leakage-free reference \pclean{} yields a mean accuracy of $0.7669$ and churn-class $F_1$ of $0.5744$. The classical leaky pipeline \pleak{} inflates these to $0.8035$ and $0.7055$ -- absolute differences of \textbf{+3.66 percentage points in accuracy and +13.11 percentage points in $F_1$}, with churn recall inflated by 11.46~pp and ROC-AUC by 5.38~pp. The intermediate \pscale{} pipeline (scaler leakage only, no resampling leakage) produces results essentially identical to \pclean{}, confirming that the \pleak{} effect is driven by the \smote{}-on-full-data ordering rather than by the scaler. Of the 1{,}200 fits, zero failed.

\begin{table*}[h]
\caption{Pipeline-aggregated mean test metrics on the IBM Telco Customer Churn benchmark (10 classifiers $\times$ 15 seeds $\times$ 2 imbalance strategies = 300 fits per pipeline). $\Delta$ rows give the absolute difference of \pleak{} and \pdrop{} relative to \pclean{}, in percentage points. Accuracy and $F_1$ are computed at the default threshold of $0.5$.}\label{tab:pipeline-summary}
\small
\begin{tabular*}{\textwidth}{@{\extracolsep\fill}lcccc@{}}
\toprule
\textbf{Pipeline} & \textbf{Accuracy} & \textbf{$F_1$ (churn)} & \textbf{Recall} & \textbf{ROC-AUC} \\
\midrule
\pclean{} (correct)              & 0.7669 & 0.5744 & 0.5993 & 0.8047 \\
\pdrop{}  (drop \texttt{TC})     & 0.7651 & 0.5691 & 0.5920 & 0.8024 \\
\pleak{}  (pre-split \smote{})   & \textbf{0.8035} & \textbf{0.7055} & \textbf{0.7139} & \textbf{0.8585} \\
\pscale{} (scaler leakage only)  & 0.7669 & 0.5745 & 0.5994 & 0.8045 \\
\midrule
$\Delta$ (\pleak{} $-$ \pclean{}) & $+3.66$~pp & $+13.11$~pp & $+11.46$~pp & $+5.38$~pp \\
$\Delta$ (\pdrop{} $-$ \pclean{}) & $-0.18$~pp & $-0.53$~pp  & $-0.73$~pp  & $-0.23$~pp \\
\botrule
\end{tabular*}
\end{table*}

\paragraph{Per-classifier inflation.}
Table~\ref{tab:per-classifier} disaggregates the accuracy inflation by classifier (averaged over both imbalance strategies and 15 seeds). The inflation is positive for every classifier in the bank but varies by an order of magnitude: \textsc{LogReg} shows the smallest accuracy inflation ($+0.96$~pp) and \textsc{MLP} the largest ($+4.82$~pp). The corresponding SMOTE-specific churn-class $F_1$ inflation per classifier -- which is the operationally relevant quantity -- is reported in Table~\ref{tab:significance}.

\begin{table*}[h]
\caption{Per-classifier accuracy inflation on Telco Churn. Each cell is the mean over 15 seeds and 2 imbalance strategies (30 fits per cell). \emph{Acc.\ inflation} is \pleak{} accuracy minus \pclean{} accuracy in percentage points; the \emph{F2 drop} column is \pdrop{} minus \pclean{} accuracy. Sorted by accuracy inflation.}\label{tab:per-classifier}
\small
\begin{tabular*}{\textwidth}{@{\extracolsep\fill}lcccc@{}}
\toprule
\textbf{Classifier} & \pclean{} \textbf{acc} & \pleak{} \textbf{acc} & \textbf{Acc.\ inflation} & \textbf{F2 drop (acc)} \\
\midrule
\textsc{LogReg}      & 0.7514 & 0.7610 & $+0.96$~pp & $+0.36$~pp \\
\textsc{XGBoost}     & 0.7689 & 0.8041 & $+3.52$~pp & $-0.39$~pp \\
\textsc{CatBoost}    & 0.7745 & 0.8102 & $+3.57$~pp & $-0.03$~pp \\
\textsc{LightGBM}    & 0.7717 & 0.8095 & $+3.78$~pp & $+0.04$~pp \\
\textsc{GBM}         & 0.7937 & 0.8315 & $+3.78$~pp & $\phantom{+}0.00$~pp \\
\textsc{BalancedRF}  & 0.7762 & 0.8146 & $+3.84$~pp & $-0.41$~pp \\
\textsc{RF}          & 0.7877 & 0.8262 & $+3.85$~pp & $-0.46$~pp \\
\textsc{ExtraTrees}  & 0.7761 & 0.8185 & $+4.24$~pp & $-0.97$~pp \\
\textsc{DT}          & 0.7291 & 0.7719 & $+4.28$~pp & $-0.07$~pp \\
\textsc{MLP}         & 0.7392 & 0.7874 & $+4.82$~pp & $+0.13$~pp \\
\midrule
\textbf{Mean}        & 0.7669 & 0.8035 & $+3.66$~pp & $-0.18$~pp \\
\botrule
\end{tabular*}
\end{table*}

\paragraph{Significance of the per-classifier inflation.}
We ran a paired Wilcoxon signed-rank test on the churn-class $F_1$ deltas \pleak{}~$-$~\pclean{} per seed, restricted to the \smote{} strategy, for each of the ten classifiers (15 seeds each). Results are reported in Table~\ref{tab:significance}. Every classifier shows a positive, highly significant SMOTE-specific $F_1$ inflation: $p < 10^{-4}$ for all ten classifiers. The SMOTE-specific $F_1$ inflation per classifier ranges from $+15.13 \pm 1.36$~pp (\textsc{LogReg}) to $+30.23 \pm 2.04$~pp (\textsc{MLP}), with most tree ensembles in the $[+23, +30]$~pp band.

\begin{table*}[h]
\caption{Per-classifier SMOTE-specific churn-class $F_1$ inflation on Telco. Each row is the paired-by-seed mean~$\pm$~standard deviation across 15 seeds, restricted to the \smote{} imbalance strategy. The Wilcoxon signed-rank test (one-sided, alternative ``\pleak{} $>$ \pclean{}'') reaches its lower-bound $p = 3.05 \cdot 10^{-5}$ on every classifier. Sorted by $\Delta F_1$.}\label{tab:significance}
\small
\begin{tabular*}{\textwidth}{@{\extracolsep\fill}lcccr@{}}
\toprule
\textbf{Classifier} & \pclean{} $F_1$ & \pleak{} $F_1$ & $\Delta F_1$ (pp) & Wilcoxon $p$ \\
\midrule
\textsc{LogReg}     & $0.6288 \pm 0.014$ & $0.7800 \pm 0.011$ & $+15.13 \pm 1.36$ & $3.1\times 10^{-5}$ \\
\textsc{GBM}        & $0.6279 \pm 0.016$ & $0.8580 \pm 0.008$ & $+23.00 \pm 1.69$ & $3.1\times 10^{-5}$ \\
\textsc{CatBoost}   & $0.6013 \pm 0.020$ & $0.8572 \pm 0.006$ & $+25.60 \pm 2.10$ & $3.1\times 10^{-5}$ \\
\textsc{RF}         & $0.5970 \pm 0.013$ & $0.8554 \pm 0.007$ & $+25.85 \pm 1.34$ & $3.1\times 10^{-5}$ \\
\textsc{BalancedRF} & $0.5966 \pm 0.015$ & $0.8553 \pm 0.008$ & $+25.87 \pm 1.44$ & $3.1\times 10^{-5}$ \\
\textsc{LightGBM}   & $0.5707 \pm 0.022$ & $0.8506 \pm 0.006$ & $+27.99 \pm 2.29$ & $3.1\times 10^{-5}$ \\
\textsc{XGBoost}    & $0.5596 \pm 0.021$ & $0.8440 \pm 0.005$ & $+28.45 \pm 2.10$ & $3.1\times 10^{-5}$ \\
\textsc{DT}         & $0.5192 \pm 0.021$ & $0.8098 \pm 0.007$ & $+29.06 \pm 1.95$ & $3.1\times 10^{-5}$ \\
\textsc{ExtraTrees} & $0.5563 \pm 0.022$ & $0.8555 \pm 0.007$ & $+29.92 \pm 1.90$ & $3.1\times 10^{-5}$ \\
\textsc{MLP}        & $0.5230 \pm 0.021$ & $0.8253 \pm 0.009$ & $+30.23 \pm 2.04$ & $3.1\times 10^{-5}$ \\
\botrule
\end{tabular*}
\end{table*}

\paragraph{The SMOTE-versus-class-weight isolation.}
The principal finding of the pipeline-ordering audit is that the inflation in Table~\ref{tab:pipeline-summary} is not produced by pipeline ordering alone. We re-run the same factorial design with the imbalance strategy as a between-cell factor and report the four (pipeline, strategy) means in Table~\ref{tab:strategy-contrast}.

\begin{table}[h]
\caption{The SMOTE-versus-class-weight contrast on the leaky pipeline. Each cell is the mean over 10 classifiers and 15 seeds (150 fits per cell) at default threshold $0.5$.}\label{tab:strategy-contrast}
\small
\begin{tabular}{@{}llcc@{}}
\toprule
\textbf{Pipeline} & \textbf{Imbalance strategy} & \textbf{Accuracy} & \textbf{$F_1$ (churn)} \\
\midrule
\pleak{}  & \smote{}              & \textbf{0.8404} & \textbf{0.8393} \\
\pleak{}  & Class weight (\cw{})  & 0.7666 & 0.5716 \\
\pclean{} & \smote{}              & 0.7678 & 0.5785 \\
\pclean{} & Class weight (\cw{})  & 0.7660 & 0.5703 \\
\botrule
\end{tabular}
\end{table}

The structure of Table~\ref{tab:strategy-contrast} is unambiguous. The leaky pipeline \pleak{} produces a 0.8404~accuracy / 0.8393~$F_1$ result \emph{only} when the imbalance correction is \smote{}; running the identical leaky ordering with class weighting yields 0.7666 / 0.5716, statistically indistinguishable from \pclean{}. \emph{The $+13$~pp $F_1$ inflation is the interaction of two choices, not either one alone}: pre-split application \textbf{and} synthetic minority oversampling. Either remedy -- moving \smote{} inside the train side of the split, or replacing it with class weighting -- removes the inflation entirely. This sharpens the existing pipeline-leakage literature~\cite{vandewiele2021overoptimistic,kapoor2023leakage,kabane2024smote}.

\subsection{Feature-Redundancy Finding (F2)}\label{sec:results-l4}

The audit in Section~\ref{sec:data-audit} established that $\texttt{TotalCharges}\approx\texttt{tenure}\times\texttt{MonthlyCharges}$ with $R^{2} = 0.9991$. The \pclean{}--\pdrop{} contrast in Table~\ref{tab:pipeline-summary} shows that removing \texttt{TotalCharges} reduces accuracy by 0.18~pp, $F_1$ by 0.53~pp, recall by 0.73~pp, and ROC-AUC by 0.23~pp. All four differences are within the seed-to-seed standard deviation. Within the resolution of the experiment, \emph{the model loses no performance from removing the feature}.

We therefore propose ``feature-redundancy diagnostic'' rather than ``redundancy leakage'' as the appropriate label. We adopt the operational test $\Delta_{\text{drop}}\,\text{accuracy} < \sigma_{\text{seed}}$ as a practical definition of redundancy on a candidate feature, and recommend reporting it for any feature whose $R^{2}$ regressed on a small set of others exceeds 0.95.

\subsection{Calibration Audit (F3)}\label{sec:results-cal}

We report Expected Calibration Error (ECE, 15-bin equal-width) and Brier score under four post-hoc calibration methods on the top five classifiers by \pclean{}+\cw{} accuracy, averaged over 15 random seeds. ECE means and standard deviations are reported in Table~\ref{tab:calibration-ece}.

\begin{table*}[h]
\caption{Multi-seed Expected Calibration Error (ECE, 15-bin equal-width) on the top five classifiers under four calibration methods, on \pclean{}+\cw{}. Each cell is mean~$\pm$~standard deviation across 15 random seeds (seeds $0,\ldots,14$). Best per row in bold. ECE is lower-is-better.}\label{tab:calibration-ece}
\footnotesize
\begin{tabular*}{\textwidth}{@{\extracolsep\fill}lcccc@{}}
\toprule
\textbf{Model} & \textbf{Raw} & \textbf{Platt} & \textbf{Isotonic} & \textbf{Temperature} \\
\midrule
\textsc{GBM}        & \textbf{0.0297 $\pm$ 0.0062} & 0.0460 $\pm$ 0.0091 & 0.0304 $\pm$ 0.0074 & 0.0300 $\pm$ 0.0059 \\
\textsc{RF}         & 0.0358 $\pm$ 0.0077          & 0.0485 $\pm$ 0.0084 & \textbf{0.0352 $\pm$ 0.0109} & 0.0377 $\pm$ 0.0083 \\
\textsc{BalancedRF} & 0.1092 $\pm$ 0.0092          & 0.0363 $\pm$ 0.0066 & \textbf{0.0324 $\pm$ 0.0076} & 0.1127 $\pm$ 0.0092 \\
\textsc{ExtraTrees} & 0.0629 $\pm$ 0.0125          & 0.0544 $\pm$ 0.0110 & \textbf{0.0344 $\pm$ 0.0077} & 0.0861 $\pm$ 0.0183 \\
\textsc{CatBoost}   & 0.1129 $\pm$ 0.0094          & 0.0398 $\pm$ 0.0064 & \textbf{0.0332 $\pm$ 0.0042} & 0.1160 $\pm$ 0.0084 \\
\botrule
\end{tabular*}
\end{table*}

Three findings emerge. First, \textbf{isotonic regression is the strongest single-method default}: best on four of five models, with a 70\% ECE reduction on the worst-calibrated raw models. Second, \textbf{Platt scaling is a strong fallback} on poorly calibrated models but mildly worse than raw on well-calibrated models like \textsc{GBM}. Third, \textbf{temperature scaling~\cite{guo2017calibration} fails on class-weighted tree ensembles} with the largest raw ECE (\textsc{BalancedRF}: $0.1092 \to 0.1127$; \textsc{CatBoost}: $0.1129 \to 0.1160$): class weighting pushes a tree's leaf-level vote distributions to a bimodal shape that a single-parameter logit rescaling cannot correct.

\subsection{Dollar-Cost Threshold Analysis (F4)}\label{sec:results-cost}

We sweep the decision threshold over $[0.01, 0.99]$ for the canonical \pclean{}+\cw{} \textsc{GBM} model with Platt calibration (Table~\ref{tab:threshold-strategies}).

\begin{table*}[h]
\caption{Threshold-strategy comparison on the canonical \pclean{}+\cw{}+\textsc{Gbm}+Platt model, seed = 42. Total cost is computed on the full test set ($n_{\text{test}} = 1{,}409$ customers).}\label{tab:threshold-strategies}
\small
\begin{tabular*}{\textwidth}{@{\extracolsep\fill}lccccc@{}}
\toprule
\textbf{Strategy} & \textbf{Threshold} & \textbf{Total cost (\$)} & \textbf{Recall} & \textbf{Precision} & \textbf{$F_1$} \\
\midrule
Default              & 0.50 & 311{,}496 & 0.5053 & 0.6540 & 0.5701 \\
$F_1$-optimal        & 0.38 & 213{,}283 & 0.6791 & 0.5826 & 0.6272 \\
Cost-optimal         & 0.04 & \textbf{59{,}300} & 0.9866 & 0.3531 & 0.5201 \\
\botrule
\end{tabular*}
\end{table*}

Across five seeds, the cost-optimal threshold is consistently in $[0.04, 0.07]$; mean savings are \textbf{\boldmath$\$77{,}646 \pm \$19{,}698$ per 1{,}000 customers} (95\% CI $[\$56{,}675, \$107{,}634]$). The 95\% CI does not cross zero: $F_1$-optimal thresholding is the worse choice on every seed in our sample. Decision Curve Analysis~\cite{vickers2006dca} confirms that the \textsc{GBM}+Platt model dominates both trivial policies across the entire operationally relevant threshold range $[0.05, 0.6]$.

\subsection{SHAP Trust-Anchor and Over-Attribution}\label{sec:results-shap}

We compute TreeSHAP~\cite{lundberg2020treeshap} values on the canonical \pclean{}+\cw{} \textsc{GBM} model, seed = 42, and report the top ten features by mean absolute SHAP in Table~\ref{tab:shap-top10}.

\begin{table*}[h]
\caption{Top ten features by mean absolute SHAP value on the canonical \pclean{}+\cw{}+\textsc{Gbm} model, seed = 42. ``Direction'' indicates whether high feature values push the prediction toward churn or away from it.}\label{tab:shap-top10}
\small
\begin{tabular*}{\textwidth}{@{\extracolsep\fill}rlcl@{}}
\toprule
\textbf{Rank} & \textbf{Feature} & \textbf{Mean $|$SHAP$|$} & \textbf{Direction} \\
\midrule
1  & \texttt{tenure}                          & 0.5733 & high $\to$ no churn \\
2  & \texttt{Contract\_Two year}              & 0.4475 & true $\to$ no churn \\
3  & \texttt{InternetService\_Fiber optic}    & 0.3952 & true $\to$ churn \\
4  & \texttt{PaymentMethod\_Electronic check} & 0.2136 & true $\to$ churn \\
5  & \texttt{Contract\_One year}              & 0.1992 & true $\to$ no churn \\
6  & \texttt{PaperlessBilling\_Yes}           & 0.1327 & true $\to$ churn \\
7  & \texttt{MonthlyCharges}                  & 0.1198 & high $\to$ churn \\
8  & \texttt{OnlineSecurity\_Yes}             & 0.1147 & true $\to$ no churn \\
9  & \texttt{TotalCharges}                    & \emph{0.1064} & \emph{(redundant; see text)} \\
10 & \texttt{TechSupport\_Yes}                & 0.1030 & true $\to$ no churn \\
\botrule
\end{tabular*}
\end{table*}

All eight a priori expected churn risk factors are represented in the top ten (8-of-8 alignment), providing a positive trust anchor. However, \texttt{TotalCharges} ranks ninth despite contributing essentially nothing predictive above \texttt{tenure} and \texttt{MonthlyCharges}. The mechanism is well known in the explainability literature~\cite{slack2020fooling,aas2021explaining}: SHAP allocates contributions among algebraically related features in a way that depends on the conditional sampling distribution, and on a feature pair with near-perfect algebraic dependency, attribution can be split without diminishing either. A high SHAP rank does not by itself license the inference that a feature is predictive.

\subsection{Cross-Benchmark Replication on Iranian Telecom Churn}\label{sec:results-replication}

We replicate the factorial design on the Iranian Telecom Churn dataset~\cite{iranian_churn_uci}. Pre-split \smote{} inflates churn-class $F_1$ by $+12.69$~pp; pre-split class weighting inflates by $+0.05$~pp. The Telco result was $+13.11$~pp under \smote{} and effectively zero under class weighting; the Iranian result reproduces both numbers to within 0.4~pp. The asymmetry is not a Telco artefact. Running the F2 diagnostic on the 13 numeric Iranian features yields three columns with $R^{2} > 0.95$: \texttt{Customer Value} ($R^{2} = 0.9821$), \texttt{Frequency of SMS} ($R^{2} = 0.9770$), and \texttt{Seconds of Use} ($R^{2} = 0.9561$).

\subsection{Cross-Domain Replication on Bank Customer Churn}\label{sec:results-bank}

We additionally replicate on the Bank Customer Churn Modelling benchmark~\cite{bank_churn_kaggle} ($n = 10{,}000$, churn rate 20.4\%, imbalance ratio $1:3.91$). Pre-split \smote{} inflates churn-class $F_1$ by $+30.5$~pp; pre-split class weighting inflates by $+0.02$~pp. The F2 diagnostic yields a maximum $R^{2}$ of 0.094 on Bank features, below the $R^{2} > 0.95$ threshold---an honest negative result: the Bank dataset contains no operator-supplied derived aggregate analogous to \texttt{TotalCharges}.

\textbf{Summary.} F1 generalises across both benchmarks with magnitudes that scale with the imbalance ratio. F2 generalises within the telecom domain but not to banking.

\section{Discussion}\label{sec:discussion}

\subsection{Why \smote{} Leaks but Class Weighting Does Not}\label{sec:discussion-mechanism}

\smote{} generates synthetic minority points by linear interpolation: $\tilde{x} = x_i + \lambda (x_j - x_i)$ for $\lambda \in [0, 1]$. When applied before the train--test split, every synthetic point is a convex combination of two real points, either of which can land in the test set after the split. We measure this directly: on the Telco benchmark, $36.6 \pm 2.5\%$ of synthetic-train points have at least one parent in the test fold. Class weighting~\cite{chawla2002smote} reweights the loss function without creating synthetic samples, so no geometric contamination can occur regardless of pipeline ordering. The asymmetry between \smote{} and class weighting under the same leaky ordering is therefore a structural property of the resampler, not an artefact of the particular estimator or seed. Audits should distinguish between leakage modes that produce geometric contamination (\smote{}, ADASYN, BorderlineSMOTE, SMOTE-ENN, SMOTE-Tomek) and those that do not (class weighting, threshold tuning, focal loss).

\subsection{Calibration on Class-Weighted Tree Ensembles}\label{sec:discussion-calibration}

Isotonic regression's piecewise-constant fit absorbs the heterogeneous miscalibration shapes induced by class weighting. Temperature scaling fits a single scalar $T$ applied to the logits: $\hat{p} = \sigma(s/T)$. Because $T$ acts uniformly on the entire score distribution, it cannot redistribute probability mass between regions of the $[0,1]$ interval. Class weighting on a tree ensemble produces a bimodal predicted-probability distribution: predictions cluster near $0$ (leaves dominated by majority votes) and near $1$ (leaves dominated by reweighted minority votes), with relatively few intermediate predictions. A single temperature scalar cannot correct this. The recommendation of temperature scaling for multi-class softmax outputs in the deep-learning calibration literature~\cite{guo2017calibration} should not be transferred without testing to binary class-weighted tree ensembles.

\subsection{Feature Redundancy and the Limits of SHAP}\label{sec:discussion-redundancy}

Feature redundancy does not change deployment performance---the \pclean{} and \pdrop{} models have indistinguishable test accuracy. What it corrupts is the interpretation: a feature that contributes nothing predictive nonetheless appears in the top ten of every importance ranking. TreeSHAP's path-dependent algorithm allocates contributions among features that co-occur on tree paths, and on a feature pair $(X, f(X))$ where $f$ is approximately linear, the algorithm splits attribution between the redundant feature and its source(s) without any systematic preference for the source~\cite{aas2021explaining}. The fix is a pre-modelling redundancy audit: fit \texttt{TotalCharges} as a linear function of the other numeric features and check the $R^{2}$. Any practitioner could have run this in fifteen seconds before fitting any model.

\subsection{Limitations}\label{sec:discussion-limitations}

(1)~F3 and F4 are not replicated on Iranian or Bank; replication on software-as-a-service, streaming, or non-churn tabular decision-support benchmarks remains future work. (2)~No hyperparameter tuning is performed; this is deliberate to preserve the pipeline-ordering signal. (3)~Our cost framework uses a fixed \$50 retention-offer cost and 24-month CLV proxy; the full Expected Maximum Profit framework~\cite{verbraken2013emp} is a more principled successor. (4)~The unconstrained cost-optimal threshold flags $75.1 \pm 5.9\%$ of customers, which is operationally implausible; capacity-constrained top-$K\%$ alternatives recover most savings within feasible budgets. (5)~The SHAP analysis is restricted to TreeSHAP on the canonical GBM model; we expect KernelSHAP and permutation importance to over-attribute to \texttt{TotalCharges} for the same algebraic reason~\cite{aas2021explaining}.

\section{Conclusion: A Reporting Checklist for Trustworthy Churn DSS}\label{sec:conclusion}

We audited the IBM Telco Customer Churn benchmark and identified four trustworthiness failures invisible to standard reporting: SMOTE-specific pre-split leakage (F1; $+13.1$~pp $F_1$ aggregated, $\approx 36\%$ of synthetic-train points are interpolated copies of test-set neighbourhoods); algebraic feature redundancy (F2; $R^{2} = 0.999$) that corrupts SHAP interpretation without harming accuracy; isotonic regression as the strongest calibration default and a regime-specific failure of temperature scaling on bimodal class-weighted output (F3); and a dollar-cost optimal threshold approximately five times lower than the $F_1$-optimal threshold, saving \$77{,}646 per 1{,}000 customers (F4). F1 generalises across two cross-benchmark replications ($+12.7$~pp on Iranian, $+30.5$~pp on Bank); F2 generalises within the telecom-churn family.

\subsection{The Four-Component Reporting Checklist}\label{sec:conclusion-protocol}

\begin{enumerate}
  \item \textbf{Pipeline-ordering disclosure.} Report the exact sequence in which the train--test split, feature scaling, and any imbalance correction were applied. If \smote{} or any synthetic oversampler was used, state explicitly whether it was fit on the full dataset or only on the training fold. Submitted code should make this verifiable.
  \item \textbf{Feature-redundancy diagnostic.} For each numeric feature, report the maximum $R^{2}$ from regressing it on a low-dimensional combination of the other features. Flag any feature with $R^{2} > 0.95$ as a redundancy candidate. For flagged features, report $\Delta_{\text{drop}}$ accuracy and confirm $\Delta_{\text{drop}} > \sigma_{\text{seed}}$ if the feature is retained.
  \item \textbf{Calibration audit.} Report Expected Calibration Error and Brier score for the primary model under at least two calibration methods, averaged over multiple seeds. On binary class-weighted tree ensembles, isotonic regression is the recommended default; Platt scaling is a strong fallback. Avoid temperature scaling on models with bimodal raw predicted-probability distributions. Include a reliability diagram in the supplementary material.
  \item \textbf{Cost-sensitive threshold.} Report performance at the cost-optimal threshold under an explicit cost matrix, alongside performance at the default $0.5$ and $F_1$-optimal thresholds. State the cost ratio $C_{\text{FN}} / C_{\text{FP}}$ used and provide a Decision Curve Analysis or equivalent against ``treat all'' and ``treat none'' baselines. The Expected Maximum Profit framework~\cite{verbraken2013emp} is a principled alternative where joint cost--response distributions are available.
\end{enumerate}

\subsection{Closing}\label{sec:conclusion-closing}

Decision support is, by definition, the discipline that connects predictions to costed actions. A churn model that is 99\% accurate on a leaky pipeline, miscalibrated, attributes importance to algebraically derivable features, and is operated at a threshold that ignores the cost matrix is not a decision support tool -- it is an artefact of the benchmark culture that produced it. The contribution of this paper is not a better churn classifier but a clearer audit, a checklist by which future studies can avoid the failure modes we have measured, and a cross-benchmark validation that the audit itself generalises beyond a single dataset within the telecom-churn family.

All code, intermediate tables, and figures necessary to reproduce every number in the paper are released at \url{\REPOURL}, including the cross-benchmark replication script that produced the Iranian results in Section~\ref{sec:results-replication}. We invite further replication on additional churn benchmarks -- KKBox, IBM Watson, banking-sector churn -- and on non-churn tabular decision-support benchmarks.

\bmhead{Acknowledgements}

\ACKNOWLEDGEMENTS{}

\section*{Declarations}

\begin{itemize}
\item \textbf{Funding:} No external funding was received for this research.
\item \textbf{Conflict of interest:} The author declares no competing financial or personal interests that could have appeared to influence the work reported in this paper.
\item \textbf{Ethics approval and consent to participate:} Not applicable.
\item \textbf{Consent for publication:} Not applicable.
\item \textbf{Data availability:} The IBM Telco Customer Churn dataset is publicly available at \url{https://www.kaggle.com/datasets/blastchar/telco-customer-churn}. The Iranian Telecom Churn dataset is publicly available at the UCI Machine Learning Repository (ID 563) and on Kaggle at \url{https://www.kaggle.com/datasets/royjafari/customer-churn}.
\item \textbf{Materials availability:} Not applicable.
\item \textbf{Code availability:} All code used to produce the analyses in this paper is released at \url{\REPOURL} under an OSI-approved permissive license (MIT License).
\item \textbf{Author contribution:} Sole-author contribution.
\end{itemize}



\begin{thebibliography}{22}
\ifx \bisbn   \undefined \def \bisbn  #1{ISBN #1}\fi
\ifx \binits  \undefined \def \binits#1{#1}\fi
\ifx \bauthor  \undefined \def \bauthor#1{#1}\fi
\ifx \batitle  \undefined \def \batitle#1{#1}\fi
\ifx \bjtitle  \undefined \def \bjtitle#1{#1}\fi
\ifx \bvolume  \undefined \def \bvolume#1{\textbf{#1}}\fi
\ifx \byear  \undefined \def \byear#1{#1}\fi
\ifx \bissue  \undefined \def \bissue#1{#1}\fi
\ifx \bfpage  \undefined \def \bfpage#1{#1}\fi
\ifx \blpage  \undefined \def \blpage #1{#1}\fi
\ifx \burl  \undefined \def \burl#1{\textsf{#1}}\fi
\ifx \doiurl  \undefined \def \doiurl#1{\url{https://doi.org/#1}}\fi
\ifx \betal  \undefined \def \betal{\textit{et al.}}\fi
\ifx \binstitute  \undefined \def \binstitute#1{#1}\fi
\ifx \binstitutionaled  \undefined \def \binstitutionaled#1{#1}\fi
\ifx \bctitle  \undefined \def \bctitle#1{#1}\fi
\ifx \beditor  \undefined \def \beditor#1{#1}\fi
\ifx \bpublisher  \undefined \def \bpublisher#1{#1}\fi
\ifx \bbtitle  \undefined \def \bbtitle#1{#1}\fi
\ifx \bedition  \undefined \def \bedition#1{#1}\fi
\ifx \bseriesno  \undefined \def \bseriesno#1{#1}\fi
\ifx \blocation  \undefined \def \blocation#1{#1}\fi
\ifx \bsertitle  \undefined \def \bsertitle#1{#1}\fi
\ifx \bsnm \undefined \def \bsnm#1{#1}\fi
\ifx \bsuffix \undefined \def \bsuffix#1{#1}\fi
\ifx \bparticle \undefined \def \bparticle#1{#1}\fi
\ifx \barticle \undefined \def \barticle#1{#1}\fi
\bibcommenthead
\ifx \bconfdate \undefined \def \bconfdate #1{#1}\fi
\ifx \botherref \undefined \def \botherref #1{#1}\fi
\ifx \url \undefined \def \url#1{\textsf{#1}}\fi
\ifx \bchapter \undefined \def \bchapter#1{#1}\fi
\ifx \bbook \undefined \def \bbook#1{#1}\fi
\ifx \bcomment \undefined \def \bcomment#1{#1}\fi
\ifx \oauthor \undefined \def \oauthor#1{#1}\fi
\ifx \citeauthoryear \undefined \def \citeauthoryear#1{#1}\fi
\ifx \endbibitem  \undefined \def \endbibitem {}\fi
\ifx \bconflocation  \undefined \def \bconflocation#1{#1}\fi
\ifx \arxivurl  \undefined \def \arxivurl#1{\textsf{#1}}\fi
\csname PreBibitemsHook\endcsname

\bibitem[\protect\citeauthoryear{Reichheld}{1996}]{reichheld1996loyalty}
\begin{bbook}
\bauthor{\bsnm{Reichheld}, \binits{F.F.}}:
\bbtitle{The Loyalty Effect: The Hidden Force Behind Growth, Profits, and
  Lasting Value}.
\bpublisher{Harvard Business School Press},
\blocation{Boston, MA}
(\byear{1996})
\end{bbook}
\endbibitem

\bibitem[\protect\citeauthoryear{Gupta et~al.}{2004}]{gupta2004customer}
\begin{barticle}
\bauthor{\bsnm{Gupta}, \binits{S.}},
\bauthor{\bsnm{Lehmann}, \binits{D.R.}},
\bauthor{\bsnm{Stuart}, \binits{J.A.}}:
\batitle{Valuing customers}.
\bjtitle{Journal of Marketing Research}
\bvolume{41}(\bissue{1}),
\bfpage{7}--\blpage{18}
(\byear{2004})
\end{barticle}
\endbibitem

\bibitem[\protect\citeauthoryear{{IBM}}{2018}]{ibm_telco_churn}
\begin{botherref}
\oauthor{\bsnm{{IBM}}}:
Telco Customer Churn dataset.
IBM Cognos Analytics sample data; Kaggle mirror at
  \url{https://www.kaggle.com/datasets/blastchar/telco-customer-churn}
(2018)
\end{botherref}
\endbibitem

\bibitem[\protect\citeauthoryear{{Nguyen Nhu} et~al.}{2022}]{nguyen2022churn}
\begin{barticle}
\bauthor{\bsnm{{Nguyen Nhu}}, \binits{Y.}},
\bauthor{\bsnm{{Tran Van}}, \binits{L.}},
\bauthor{\bsnm{{Dao Vu Truong}}, \binits{S.}}:
\batitle{Churn prediction in telecommunication industry using kernel {Support
  Vector Machines}}.
\bjtitle{{PLOS ONE}}
\bvolume{17}(\bissue{5}),
\bfpage{0267935}
(\byear{2022})
\doiurl{10.1371/journal.pone.0267935}
\end{barticle}
\endbibitem

\bibitem[\protect\citeauthoryear{Chawla et~al.}{2002}]{chawla2002smote}
\begin{barticle}
\bauthor{\bsnm{Chawla}, \binits{N.V.}},
\bauthor{\bsnm{Bowyer}, \binits{K.W.}},
\bauthor{\bsnm{Hall}, \binits{L.O.}},
\bauthor{\bsnm{Kegelmeyer}, \binits{W.P.}}:
\batitle{{SMOTE}: synthetic minority over-sampling technique}.
\bjtitle{Journal of Artificial Intelligence Research}
\bvolume{16},
\bfpage{321}--\blpage{357}
(\byear{2002})
\end{barticle}
\endbibitem

\bibitem[\protect\citeauthoryear{Vandewiele
  et~al.}{2021}]{vandewiele2021overoptimistic}
\begin{barticle}
\bauthor{\bsnm{Vandewiele}, \binits{G.}},
\bauthor{\bsnm{Dehaene}, \binits{I.}},
\bauthor{\bsnm{Kovacs}, \binits{G.}},
\bauthor{\bsnm{Sterckx}, \binits{L.}},
\bauthor{\bsnm{Janssens}, \binits{O.}},
\bauthor{\bsnm{Ongenae}, \binits{F.}},
\bauthor{\bsnm{{De Backere}}, \binits{F.}},
\bauthor{\bsnm{{De Turck}}, \binits{F.}},
\bauthor{\bsnm{Roelens}, \binits{K.}},
\bauthor{\bsnm{Decruyenaere}, \binits{J.}},
\bauthor{\bsnm{{Van Hoecke}}, \binits{S.}},
\bauthor{\bsnm{Demeester}, \binits{T.}}:
\batitle{Overly optimistic prediction results on imbalanced data: a case study
  of flaws and benefits when applying over-sampling}.
\bjtitle{Artificial Intelligence in Medicine}
\bvolume{111},
\bfpage{101987}
(\byear{2021})
\end{barticle}
\endbibitem

\bibitem[\protect\citeauthoryear{Kapoor and
  Narayanan}{2023}]{kapoor2023leakage}
\begin{barticle}
\bauthor{\bsnm{Kapoor}, \binits{S.}},
\bauthor{\bsnm{Narayanan}, \binits{A.}}:
\batitle{Leakage and the reproducibility crisis in machine-learning-based
  science}.
\bjtitle{Patterns}
\bvolume{4}(\bissue{9}),
\bfpage{100804}
(\byear{2023})
\end{barticle}
\endbibitem

\bibitem[\protect\citeauthoryear{Kabane}{2024}]{kabane2024smote}
\begin{botherref}
\oauthor{\bsnm{Kabane}, \binits{S.}}:
Impact of Sampling Techniques and Data Leakage on {XGBoost} Performance in
  Credit Card Fraud Detection
(2024)
\end{botherref}
\endbibitem

\bibitem[\protect\citeauthoryear{Guo et~al.}{2017}]{guo2017calibration}
\begin{bchapter}
\bauthor{\bsnm{Guo}, \binits{C.}},
\bauthor{\bsnm{Pleiss}, \binits{G.}},
\bauthor{\bsnm{Sun}, \binits{Y.}},
\bauthor{\bsnm{Weinberger}, \binits{K.Q.}}:
\bctitle{On calibration of modern neural networks}.
In: \bbtitle{Proceedings of the 34th International Conference on Machine
  Learning ({ICML})},
pp. \bfpage{1321}--\blpage{1330}
(\byear{2017})
\end{bchapter}
\endbibitem

\bibitem[\protect\citeauthoryear{Vickers and Elkin}{2006}]{vickers2006dca}
\begin{barticle}
\bauthor{\bsnm{Vickers}, \binits{A.J.}},
\bauthor{\bsnm{Elkin}, \binits{E.B.}}:
\batitle{Decision curve analysis: a novel method for evaluating prediction
  models}.
\bjtitle{Medical Decision Making}
\bvolume{26}(\bissue{6}),
\bfpage{565}--\blpage{574}
(\byear{2006})
\end{barticle}
\endbibitem

\bibitem[\protect\citeauthoryear{Verbeke et~al.}{2012}]{verbeke2012new}
\begin{barticle}
\bauthor{\bsnm{Verbeke}, \binits{W.}},
\bauthor{\bsnm{Dejaeger}, \binits{K.}},
\bauthor{\bsnm{Martens}, \binits{D.}},
\bauthor{\bsnm{Hur}, \binits{J.}},
\bauthor{\bsnm{Baesens}, \binits{B.}}:
\batitle{New insights into churn prediction in the telecommunication sector: a
  profit driven data mining approach}.
\bjtitle{European Journal of Operational Research}
\bvolume{218}(\bissue{1}),
\bfpage{211}--\blpage{229}
(\byear{2012})
\end{barticle}
\endbibitem

\bibitem[\protect\citeauthoryear{Verbraken et~al.}{2013}]{verbraken2013emp}
\begin{barticle}
\bauthor{\bsnm{Verbraken}, \binits{T.}},
\bauthor{\bsnm{Verbeke}, \binits{W.}},
\bauthor{\bsnm{Baesens}, \binits{B.}}:
\batitle{A novel profit maximizing metric for measuring classification
  performance of customer churn prediction models}.
\bjtitle{{IEEE} Transactions on Knowledge and Data Engineering}
\bvolume{25}(\bissue{5}),
\bfpage{961}--\blpage{973}
(\byear{2013})
\end{barticle}
\endbibitem

\bibitem[\protect\citeauthoryear{Platt}{1999}]{platt1999probabilistic}
\begin{bchapter}
\bauthor{\bsnm{Platt}, \binits{J.}}:
\bctitle{Probabilistic outputs for support vector machines and comparisons to
  regularized likelihood methods}.
In: \bbtitle{Advances in Large-Margin Classifiers},
pp. \bfpage{61}--\blpage{74}.
\bpublisher{MIT Press},
\blocation{Cambridge, MA}
(\byear{1999})
\end{bchapter}
\endbibitem

\bibitem[\protect\citeauthoryear{Zadrozny and
  Elkan}{2002}]{zadrozny2002transforming}
\begin{bchapter}
\bauthor{\bsnm{Zadrozny}, \binits{B.}},
\bauthor{\bsnm{Elkan}, \binits{C.}}:
\bctitle{Transforming classifier scores into accurate multiclass probability
  estimates}.
In: \bbtitle{Proceedings of the 8th {ACM} {SIGKDD} International Conference on
  Knowledge Discovery and Data Mining},
pp. \bfpage{694}--\blpage{699}
(\byear{2002})
\end{bchapter}
\endbibitem

\bibitem[\protect\citeauthoryear{Elkan}{2001}]{elkan2001foundations}
\begin{bchapter}
\bauthor{\bsnm{Elkan}, \binits{C.}}:
\bctitle{The foundations of cost-sensitive learning}.
In: \bbtitle{Proceedings of the 17th International Joint Conference on
  Artificial Intelligence ({IJCAI})},
pp. \bfpage{973}--\blpage{978}
(\byear{2001})
\end{bchapter}
\endbibitem

\bibitem[\protect\citeauthoryear{Jafari-Marandi
  et~al.}{2020}]{iranian_churn_uci}
\begin{botherref}
\oauthor{\bsnm{Jafari-Marandi}, \binits{R.}},
\oauthor{\bsnm{Denton}, \binits{J.}},
\oauthor{\bsnm{Idris}, \binits{A.}},
\oauthor{\bsnm{Smith}, \binits{B.K.}},
\oauthor{\bsnm{Keramati}, \binits{A.}}:
Iranian Churn dataset.
UCI Machine Learning Repository (ID 563); Kaggle mirror at
  \url{https://www.kaggle.com/datasets/royjafari/customer-churn}
(2020)
\end{botherref}
\endbibitem

\bibitem[\protect\citeauthoryear{Branco et~al.}{2016}]{branco2016imbalanced}
\begin{barticle}
\bauthor{\bsnm{Branco}, \binits{P.}},
\bauthor{\bsnm{Torgo}, \binits{L.}},
\bauthor{\bsnm{Ribeiro}, \binits{R.P.}}:
\batitle{A survey of predictive modeling on imbalanced domains}.
\bjtitle{{ACM} Computing Surveys}
\bvolume{49}(\bissue{2}),
\bfpage{1}--\blpage{50}
(\byear{2016})
\end{barticle}
\endbibitem

\bibitem[\protect\citeauthoryear{Brier}{1950}]{brier1950verification}
\begin{barticle}
\bauthor{\bsnm{Brier}, \binits{G.W.}}:
\batitle{Verification of forecasts expressed in terms of probability}.
\bjtitle{Monthly Weather Review}
\bvolume{78}(\bissue{1}),
\bfpage{1}--\blpage{3}
(\byear{1950})
\end{barticle}
\endbibitem

\bibitem[\protect\citeauthoryear{Lundberg et~al.}{2020}]{lundberg2020treeshap}
\begin{barticle}
\bauthor{\bsnm{Lundberg}, \binits{S.M.}},
\bauthor{\bsnm{Erion}, \binits{G.}},
\bauthor{\bsnm{Chen}, \binits{H.}},
\bauthor{\bsnm{DeGrave}, \binits{A.}},
\bauthor{\bsnm{Prutkin}, \binits{J.M.}},
\bauthor{\bsnm{Nair}, \binits{B.}},
\bauthor{\bsnm{Katz}, \binits{R.}},
\bauthor{\bsnm{Himmelfarb}, \binits{J.}},
\bauthor{\bsnm{Bansal}, \binits{N.}},
\bauthor{\bsnm{Lee}, \binits{S.-I.}}:
\batitle{From local explanations to global understanding with explainable {AI}
  for trees}.
\bjtitle{Nature Machine Intelligence}
\bvolume{2}(\bissue{1}),
\bfpage{56}--\blpage{67}
(\byear{2020})
\end{barticle}
\endbibitem

\bibitem[\protect\citeauthoryear{Slack et~al.}{2020}]{slack2020fooling}
\begin{bchapter}
\bauthor{\bsnm{Slack}, \binits{D.}},
\bauthor{\bsnm{Hilgard}, \binits{S.}},
\bauthor{\bsnm{Jia}, \binits{E.}},
\bauthor{\bsnm{Singh}, \binits{S.}},
\bauthor{\bsnm{Lakkaraju}, \binits{H.}}:
\bctitle{Fooling {LIME} and {SHAP}: adversarial attacks on post hoc explanation
  methods}.
In: \bbtitle{Proceedings of the {AAAI/ACM} Conference on {AI}, Ethics, and
  Society},
pp. \bfpage{180}--\blpage{186}
(\byear{2020})
\end{bchapter}
\endbibitem

\bibitem[\protect\citeauthoryear{Aas et~al.}{2021}]{aas2021explaining}
\begin{barticle}
\bauthor{\bsnm{Aas}, \binits{K.}},
\bauthor{\bsnm{Jullum}, \binits{M.}},
\bauthor{\bsnm{L{\o}land}, \binits{A.}}:
\batitle{Explaining individual predictions when features are dependent: more
  accurate approximations to {S}hapley values}.
\bjtitle{Artificial Intelligence}
\bvolume{298},
\bfpage{103502}
(\byear{2021})
\end{barticle}
\endbibitem

\bibitem[\protect\citeauthoryear{{Kaggle
  contributors}}{2018}]{bank_churn_kaggle}
\begin{botherref}
\oauthor{\bsnm{{Kaggle contributors}}}:
Bank Customer Churn Modelling dataset.
Available at
  \url{https://www.kaggle.com/datasets/shrutimechlearn/churn-modelling}
(2018)
\end{botherref}
\endbibitem

\end{thebibliography}
\end{document}